\documentclass[letterpaper, 9pt, conference]{ieeeconf}  

\usepackage{graphicx}
\usepackage{amsmath}
\usepackage{amssymb}
\usepackage{booktabs}
\usepackage{multirow}
\usepackage{tabularx}
\usepackage{bbding}

\newcommand{\eg}{\textit{e}.\textit{g}.}

\newcommand{\ie}{\textit{i}.\textit{e}.}

\IEEEoverridecommandlockouts                              

\usepackage{graphics} 
\usepackage{epsfig} 
\usepackage{mathptmx} 
\usepackage{times} 
\usepackage{amsmath} 
\usepackage{amssymb}  
\usepackage{hyperref}
\hypersetup{pdfstartview=FitH,
            colorlinks=true,
            linkcolor=red,
            anchorcolor=blue,
            citecolor=blue
            }
\title{\LARGE \bf
SRFNet: Monocular Depth Estimation with Fine-grained Structure via Spatial Reliability-oriented Fusion
of Frames and Events}

\author{Tianbo Pan$^{1*}$, Zidong Cao$^{1*}$, \textit{IEEE Student Member}, Lin Wang$^{1,2}$$^\dagger$, \textit{IEEE Member} 
\thanks{$^\dagger$Corresponding author, $^*$ Co-first authors.}
\thanks{$^{1}$ T. Pan is with AI Thrust,
        HKUST(GZ), Guangzhou, China,
        Email: {\tt\small tpan695@connect.hkust-gz.edu.cn}}%
\thanks{$^{1}$Z. Cao is with AI Thrust,
        HKUST(GZ), Guangzhou, China,
        Email: {\tt\small caozidong1996@gmail.com}}%
\thanks{$^{1,2}$L. Wang is with AI Thrust, HKUST(GZ) and Dept. of CSE, HKUST, China,
        Email: {\tt\small linwang@ust.hk}}}

\begin{document}

\maketitle
\thispagestyle{empty}
\pagestyle{empty}

\begin{abstract}

Monocular depth estimation is a crucial task to measure distance relative to a camera, which is important for applications, such as robot navigation and self-driving.  
Traditional frame-based methods suffer from performance drops due to the limited dynamic range and motion blur. Therefore, recent works leverage novel event cameras to complement or guide the frame modality via frame-event feature fusion. However, event streams exhibit spatial sparsity, leaving some areas unperceived, especially in regions with marginal light changes. Therefore, direct fusion methods, \eg, RAMNet~\cite{Gehrig2021CombiningEA}, often ignore the contribution of the most confident regions of each modality. This leads to structural ambiguity in the modality fusion process, thus degrading the depth estimation performance.
In this paper, we propose a novel Spatial Reliability-oriented Fusion Network (SRFNet), that can estimate depth with fine-grained structure at both daytime and nighttime. Our method consists of two key technical components. Firstly, we propose an attention-based interactive fusion (AIF) module that applies spatial priors of events and frames as the initial masks and learns the consensus regions to guide the inter-modal feature fusion. 
The fused feature are then fed back to enhance the frame and event feature learning. Meanwhile, it utilizes an output head to generate a fused mask, which is iteratively updated for learning consensual spatial priors. 
Secondly, we propose the Reliability-oriented Depth Refinement (RDR) module to estimate dense depth with the fine-grained structure based on the fused features and masks. 
We evaluate the effectiveness of our method on the synthetic and real-world datasets, which shows that, even without pretraining, our method outperforms the prior methods, \eg, RAMNet~\cite{Gehrig2021CombiningEA}, especially in night scenes. Our project homepage: \url{https://vlislab22.github.io/SRFNet}.

\end{abstract}

\section{INTRODUCTION}

Monocular depth estimation is a fundamental vision task with various applications, such as robotic navigation~\cite{vidal2018ultimate} and self-driving~\cite{guizilini20203d}. 
In the last decade, deep learning-based methods have shown significant performance gains for monocular depth estimation using the standard frame-based cameras~\cite{Li2018MegaDepthLS,Gurram2021MonocularDE}.
However, these methods encounter critical challenges, particularly in extreme, \eg, high-speed motion and low-light scenes. 


\begin{figure}[t!]
    \centering
    \includegraphics[width=\linewidth]{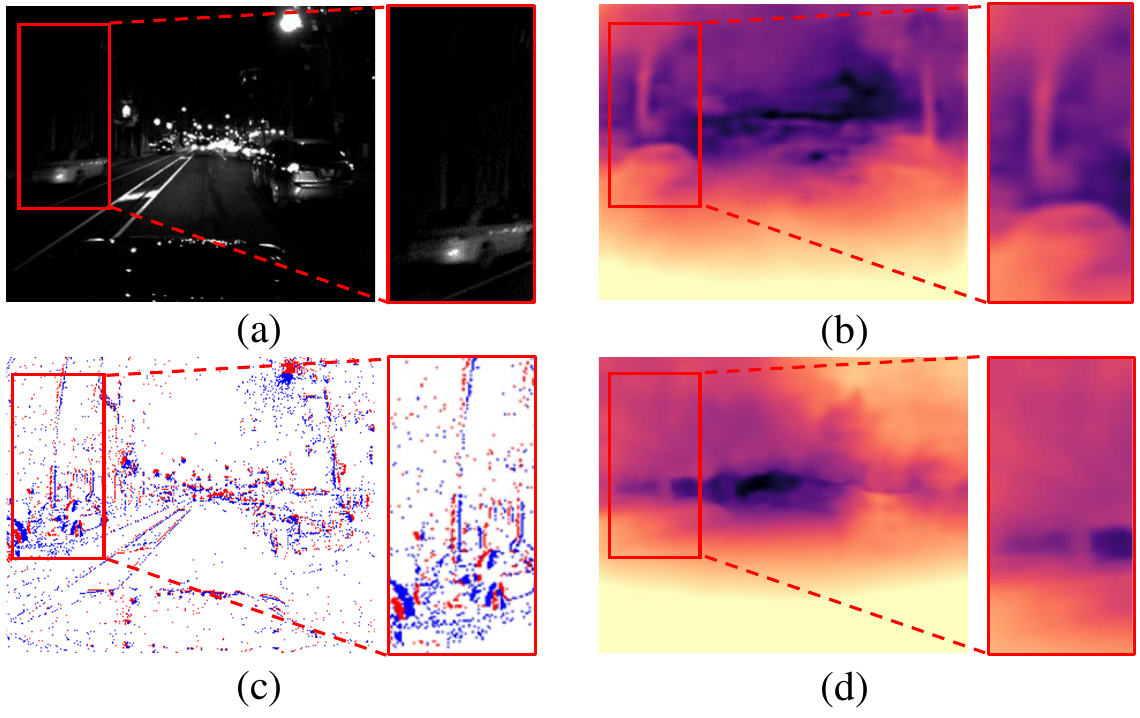}
    \caption{(a) and (c) are the input intensity frame and events, respectively; (b) is our result while (d) is the result of RAMNet~\cite{Gehrig2021CombiningEA}. Within the red rectangle, we zoom in on a challenging scenario where a car and tree in the frame are exposed under poor lighting and show discontinuous edge information in the events. Our SRFNet excels in predicting dense depth with fine-grained structural details compared with RAMNet.}
    \label{teaser fig}
    \vspace{-10pt}
\end{figure}

Recently, using novel event cameras to estimate depth has experienced a surge in popularity~\cite{zheng2023deep}. Event cameras are bio-inspired sensors that asynchronously detect the intensity changes of a scene and output event streams encoding time, pixel location, and polarity (\ie, sign for the intensity changes). Event cameras show distinct advantages over RGB-based cameras, such as high dynamic range (HDR) and high temporal resolution. 
Accordingly, some event-based methods have been proposed to directly estimate depth from event data~\cite{HidalgoCarrio2020LearningMD,wang2021dual}. Although these methods show superiority in extreme visual conditions, the intrinsic sparsity and unexpected noise of events often cause structural ambiguity of objects.

To complement the RGB cameras, recent works attempt to leverage event cameras as guidance for better depth estimation~\cite{Gehrig2021CombiningEA,Shi2023EVENAE}. These methods
learn the modal-specific features independently and fuse RGB-event features without distinguishing the contribution of the most confident regions of each modality. This leads to structural ambiguity between the two modalities, which may hamper the depth estimation performance. 
Also, although the prior work RAMNet~\cite{Gehrig2021CombiningEA} considers the temporal information of event data, it overlooks the importance of spatial sparsity, which is crucial for structural information extraction. Intuitively, it remains a challenge to effectively harness the strengths of both modalities. 

In this paper, we find that inter-modal cross-referencing can greatly benefit feature learning of each modality. 
Accordingly, we present a novel framework Spatial Reliability-oriented Fusion Network (\textbf{SRFNet}) that can estimate dense depth with fine-grained structure at both daytime and nighttime, as shown in Fig.~\ref{teaser fig}(b). Our SRFNet consists of two key technical components (See Fig.~\ref{pipeline}).
Firstly, we propose an Attention-based Interactive Fusion (AIF) module, including a series of Consensus Learning (CL) blocks to adaptively integrate the two modalities while employing the fused feature to enhance each modality's feature learning (Sec.~\ref{sec:network}). Specifically, the AIF module initially applies spatial priors of event and frame as the initial masks to learn the consensus regions of two modalities and generate a fused feature through an attention-based layer. The fused features are then fed back to enhance the frame and event feature learning. Meanwhile, it leverages an output head to generate a fused mask for iterative spatial prior learning. That enable the spatial prior progressively to have more consensus information.
Through the iterative fusion process of CL blocks, it is possible to effectively leverage the superiority of each modality.

Secondly, we propose a Reliability-oriented Depth Refine (RDR) module to estimate dense depth with the fine-grained structure based on the fused feature and masks. Concretely, we employ a decoder together with a temporal learning layer to estimate a depth feature. Then, the fused masks from CL blocks are stacked and incorporated with the depth feature. The original depth feature is used to obtain the affinity map and coarse depth map with two individual output heads, respectively, while the incorporated depth feature is used to obtain the confidence map. Finally, the affinity map, coarse depth map, and confidence map are fed into the spatial propagation network (SPN) to generate the fine-grained depth map.


In summary, the main contributions of our paper are: 
\begin{itemize}
\item We propose a novel SRFNet that can estimate dense depth with fine-grained structure at both daytime and nighttime.  
\item We propose the AIF module that interactively enhances two modalities' feature learning and the RDR module that further obtains fine-grained depth estimation.
\item We conduct extensive experiments on both synthetic and real-world datasets. We compare our method with the state-of-the-art (SOTA) frame-based, event-based, and frame-event fusion methods. The experimental results reveal that \textit{even without pre-training}, our SRFNet yields a \textbf{substantial improvement} over the SOTA methods, \eg, RAMNet~\cite{Gehrig2021CombiningEA}, and exhibits outstanding generalization capabilities, especially in night scenes.

\end{itemize}

\section{Related Works}

\noindent \textbf{Monocular Depth Estimation.}
Frame-based monocular depth estimation has been studied in both images~\cite{Alhashim2018HighQM} and videos~\cite{Yin2018GeoNetUL}. Previous approaches in this domain have primarily revolved around two key aspects: 1) Training strategy, including supervised learning~\cite{Ranftl2021VisionTF, Fu2018DeepOR}, unsupervised learning~\cite{Yin2018GeoNetUL}, and transfer learning~\cite{Alhashim2018HighQM}. 2) Backbone architectures, such as vision transformers (ViT)~\cite{Ranftl2021VisionTF}. Although frame-based methods achieve excellent performance, their performance often drops greatly in challenging scenes, such as those with poor lighting conditions and high-speed motion. 

Recently, event cameras have garnered substantial attention due to their ability to capture high dynamic range (HDR) scenes in extreme visual conditions~\cite{zheng2023deep, HidalgoCarrio2020LearningMD, Shi2023ImprovedED, Zhang2022SpikeTM, Liu2022EventbasedMD}. This has led to increased research interests in event-frame fusion for monocular depth estimation. For example,  EVEN~\cite{Shi2023EVENAE} integrates multi-modal features to enhance the frame feature learning. RAMNet~\cite{Gehrig2021CombiningEA} introduces an RNN-based network to further utilize the asynchronous property of events for better fusion. However, these methods pay less attention to the spatial reliability of each modality during fusion, treating both informative and noisy regions equally. \textit{Differently, our proposed SRFNet that can better distinguish the contributions of the most confident regions of each modality during fusion. It can estimate depth map with fine-grained structure, especially in nighttime scenes (See Fig.~\ref{teaser fig}(b)).}

\noindent \textbf{Event-Frame Fusion.}
The complementary nature of event and frame data has been extensively explored in various vision tasks, like object detection~\cite{Wang2021VisEventRO,Zhang2021ObjectTB,Zhou2022RGBEventFF,Tomy2022FusingEA}, semantic segmentation~\cite{Sun2020RealTimeFN,Zhang2020ISSAFEIS,choi2020learning,Liu2022CMXCF}, and depth estimation~\cite{Zuo2021AccurateDE,MostafaviIsfahani2021EventIntensitySE,Cho2022SelectionAC,Gehrig2021CombiningEA,Shi2023EVENAE}. The above mentioned fusion methods can be classified into two categories based on whether the interaction is considered.

The first type is fusion without interaction, where each modal feature is learned independently, and the fused feature is only utilized by the decoder. For instance, RAMNet~\cite{Gehrig2021CombiningEA} learns event and frame features asynchronously and separately. EVEN~\cite{Shi2023EVENAE} and HDES~\cite{Zuo2021AccurateDE} add multi-modal features and use the attention mechanism to further incorporate them, while~\cite{Zhang2021ObjectTB} leverages the cross-modal attention to achieve multi-modal feature integration. Also,~\cite{Tomy2022FusingEA} proposes to concatenate synchronous multi-scale multi-modal features.
The second type is fusion with interaction, which involves getting feedback from the fused result.
Among the approaches, ~\cite{Wang2021VisEventRO,Zhou2022RGBEventFF,Lu2023LearningSI} explore the feedback paradigm, in which the learned fused feature is added with the modal-specific features to enhance the subsequent feature learning.~\cite{Cho2022SelectionAC} proposes to generate a cross-similarity feature and employs convolutions to merge the similarity feature with modal-specific features. However, these works fuse frame-event features without distinguishing the contribution of the most confident regions of each modality, which might result in structural ambiguity in the extreme scenes. \textit{In contrast, our SRFNet applies spatial priors of events and frames and learns the consensus regions to guide the inter-modal feature fusion. }

\noindent \textbf{Spatial Learning for Depth Estimation.}
Spatial Propagation Network (SPN)~\cite{Liu2017LearningAV} aims to refine the coarse depth to achieve fine-grained details in the estimated depth map. For instance, CSPN++~\cite{Cheng2019CSPNLC} develops adaptable kernel sizes and iteration numbers to enhance the propagation process. The effectiveness of spatial learning is further boosted by incorporating deformable local neighbors~\cite{Park2020NonLocalSP,Xu2020DeformableSP}. 
\textit{Differently, we leverage the fused masks from the AIF module to refine the confidence map, thus giving more accurate guidance for the refinement process. }

\section{METHOD}

In this section, we first describe the data preprocessing, including the event representation and modality-specific mask initialization. We then outline the network architecture of our SRFNet framework, composed of the Attention-based Interactive Fusion (AIF) module and Reliability-oriented Depth Refine (RDR) module. We now describe the details.  

\subsection{Data preprocessing}
\textbf{Event Representation.} 
An event $e=(x,y,t,p)$ is generated when the log intensity difference at a pixel location ($x, y$) and time $t$ exceeds a threshold $C$. The polarity $p = \pm1$ indicates the intensity change direction. 
Following~\cite{HidalgoCarrio2020LearningMD}, we accumulate events using non-overlapping event windows, spanning a fixed interval $\Delta T$. The  $k_{th}$ event window is represented into a frame-like representation $\epsilon_{k}=\{e_i\}_{i\in[0,\mathbf{N-1}]}$, where $N$ is number of event in the window and the fixed time interval $\begin{aligned}\Delta T=t_{N-1}^{k}-t_{0}^{k}\end{aligned}$.
encoded as spatio-temporal voxel grids $V_{k}$ with dimensions $B \times H \times W$. Here, $B$ denotes the number of temporal bins. Following~\cite{HidalgoCarrio2020LearningMD,Gehrig2021CombiningEA}, we set $B$ to 5 and normalize the voxel grid representation. The resulting volume $V_{k}$ forms a frame-like tensor, making it compatible with processing using convolutional neural networks.

\begin{figure}[t!]
    \centering
    \includegraphics[width=0.95\linewidth]{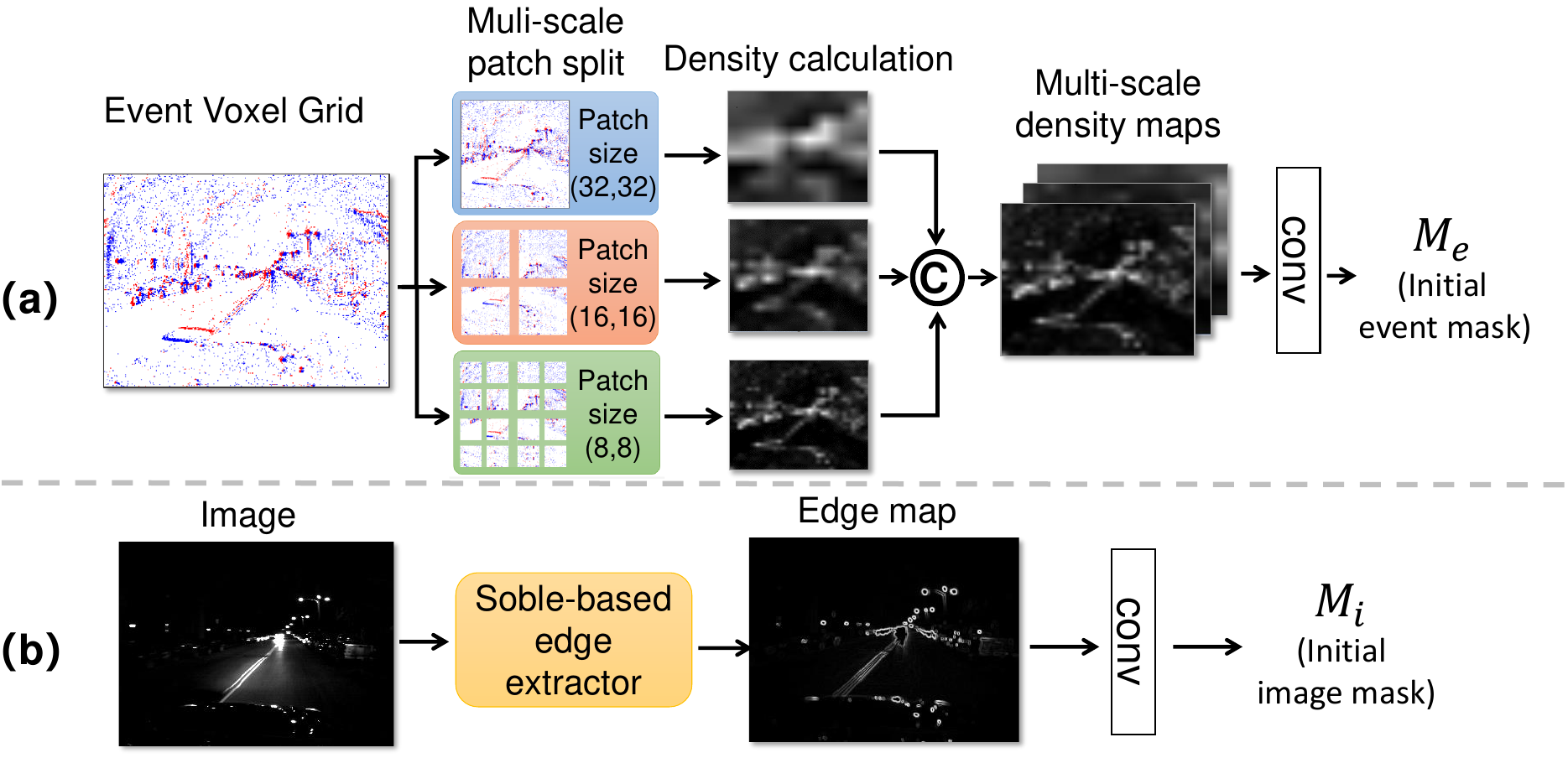}
    \caption{Illustration of the initialization of modal-specific masks, $M_{i}$ for the frame and $M_{e}$ for the events.}
    \label{Mask initialization}
\vspace{-10pt}
\end{figure}

\textbf{Modal-specific Mask Initialization.} 
In order to get more effective information extraction
as ~\cite{Sun2021EventBasedFF}, we design learnable modality-specific masks that allow each modality to contribute its most confident regions.  Instead of deriving masks from extracted features, we initialize them according to each modality's inherent properties and develop separate learning for iteratively updating modality-specific masks.  

The mask serves as a confidence map to indicate the informative content of various regions. For initializing the event masks, each time bin of the voxel grid is split into multi-scale non-overlapping patches (\ie, $8\times8$, $16\times16$, $32\times32$), as depicted in Fig.~\ref{Mask initialization}(a). The event density within each patch is calculated and then normalized by dividing it by the average density of all patches.  We average the density maps from different time bins under the same patch size and concatenate the averaged density maps from multi-scale patches. 
These concatenated density maps are subsequently input into a $3\times3$ convolution to obtain an event mask $M_{e}$. 
Fig.~\ref{Mask initialization}(b) depicts the initialization of the frame mask, which is based on the assumption that regions with more edge information are more likely to have significant changes in depth~\cite{Li2021EdgeAwareMD,Pu2021RINDNetED}. Our method employs the Sobel edge extractor~\cite{sobel19683x3} and feeds the edge map to convolution to learn to the frame mask $M^k_{i}$. These two masks are taken as the inputs to the SRFNet.

\subsection{Network Architecture} 
\label{sec:network}

\begin{figure*}[t!]
    \centering
    \includegraphics[width=0.95\textwidth]{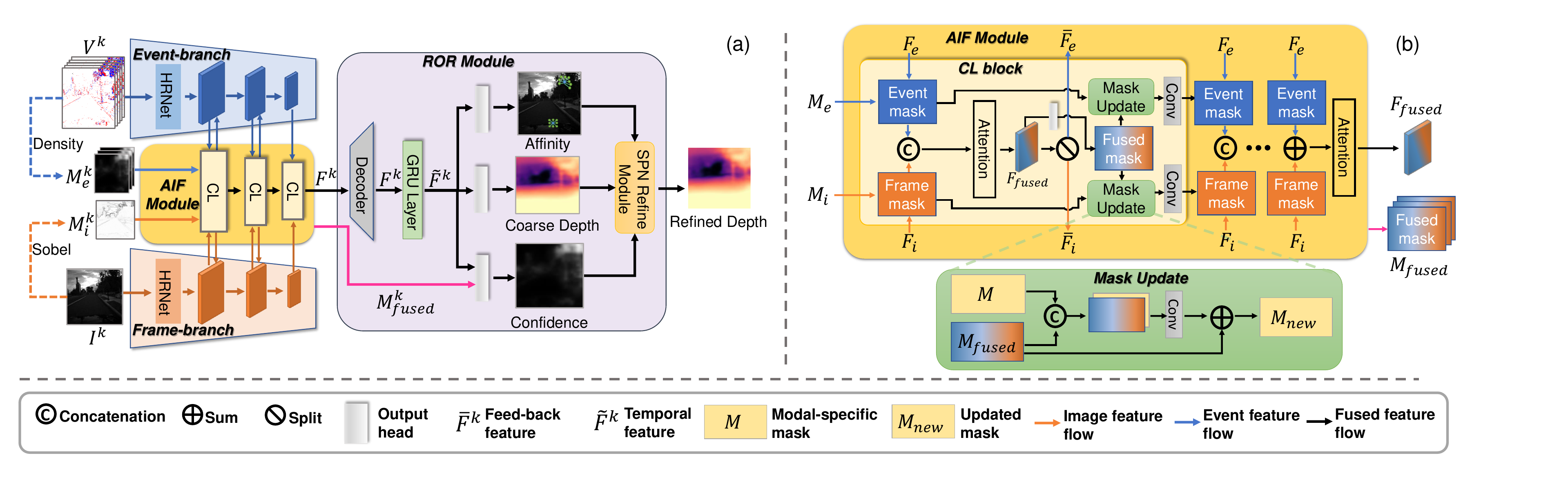}
    \vspace{-10pt}
    \caption{(a) is the overview of our proposed framework, and (b) depicts the details of the AIF module. }
    \label{pipeline}
    \vspace{-6pt}
\end{figure*}

As shown in Fig.~\ref{pipeline}, our SRFNet predicts dense log depth maps with fine-grained structure from preprocessed event voxel grids and frame sequence. For the $k_{th}$ paired input: event voxel grids $V^k$, frames $I^k$, and event mask $M^k_e$ and frame mask $M^k_i$, the first Attention-based Interactive Fusion(AIF) module generate  fused features $F^k$ and stacked fused masks $M^k_fu$, which contains consensual information learned from modalities. 
The fused features and masks are then passed to the second Reliability-oriented Depth Refine (RDR) module, which functions as a decoder, generating an initial coarse prediction and further refining its structural details.

\noindent \textbf{Attention-based Interactively Fusion (AIF) Module}. It receives preprocessed modal-specific masks and pyramid modal features from event and frame branches as input. It utilizes spatial priors to guide inter-modal fusion, promoting consensual fused feature learning and enhancing modal-specific feature learning.
 
Specifically, the event and frame branch extend the encoder design in~\cite{Gurram2021MonocularDE}. The High-Resolution Network (HRNet)~\cite{Wang2019DeepHR} is chosen as the encoder head due to its proven excellence in perceiving object layouts. $F_{e}^{k}, F_{i}^{k}$ represents the event and frame feature extracted from the $k_{th}$ frame and voxel grid  $I^{k}, V^{k}$, respectively.
Additionally, to align with the resolution of learned feature, the modal-specific masks $M_{e}^{k}, M_{i}^{k}$ are downsampled using a convolution with a $3\times3$ kernel, a stride of 2, and padding of 1. 

Our AIF module leverage multiple Consensus Learning (CL) blocks to facilitate the interactive fusion, as illustrated in Fig.~\ref{pipeline}.
The CL block, depicted in Fig.~\ref{pipeline}(b), is an attention-based fusion block that generates fused features and masks.
Prior to the combination, CL block emphasizes the modal-specific features $F_{e},F_{i} \in R^{H\times W\times C}$ , with their masks $M_{e},M_{i} \in R^{H\times W\times 1}$, which represent spatial priors denoting pixel-wise reliability. The emphasized features are then concatenated and passed through an attention layer to learn a fused feature $F_{fused} \in R^{H\times W \times 2C}$. 
The fused feature $F_{fused}$ is then split to gain feedback features $\overline{F}_{e},\overline{F}_{i}$, and then add them to $F_{e}^{k}, F_{i}^{k}$, enhancing the subsequent event and frame feature learning.
Meanwhile, the fused feature leverages an output head to generate a fused mask containing consensual information. Subsequently, the fused mask updates the modal-specific mask with a residual connection manner, as shown in Fig.~\ref{pipeline}(b).  The event and frame masks are iteratively updated through the CL blocks to incorporate more consensual information. In the last CL block, the emphasized $F_{e}^{k}, F_{i}^{k}$ are directly added and passed to attention layer to learn the final fused feature and mask. Ultimately, the AIF module stacks all masks generated by the CL blocks and then passes the fused features $F^k$ and stacked masks $M^k_{fused}$ to next module.


\noindent \textbf{Reliability-oriented Depth Refinement (RDR) Module.} 
\label{RDR}
Our Reliability-oriented Depth Refinement (RDR) module serves as an aggregation of decoder and refinement block. It enhances the accuracy of coarse predictions by incorporating a temporal layer that strengthens the fused feature with temporal correlations. Additionally, the RDR module optimizes the generation of affinities, significantly improving the performance of the NLSPN-based refinement block, resulting in fine-grained depth structures and reduced artifacts.

Inspired by ~\cite{Zhang2019ExploitingTC}, we optimize the decoder structure in ~\cite{Godard2018DiggingIS} by adding a temporal layer near the output head, which learns temporal correlations. The temporal layer is replaceable, we use convolution Gated Recurrent Unit (convGRU)~\cite{Siam2016ConvolutionalGR} in our SRFNet. The convGRU layer utilizes an updatable latent state $S$ to capture temporal information. By integrating fused feature $F^k$ and previous state $S^{k-1}$ , it yields both the temporal feature $\widetilde{F}^k$ and current state 
$S^k$, as described in the following formulation:
\begin{equation}
    (\widetilde{F}^k_{fused},S^k)=\textit{{convGRU}}(F^k_{fused},S^{k-1})
\end{equation}
Subsequently, the $\widetilde{F}^k_{fused}$ is utilized to generate the coarse depth prediction via a depth output head, while $S^k$ spreads the temporal information for $k+1_{th}$ prediction.

To further eliminate the ambiguity of  structural details of depth map, the RDR module incorporates the Non-Local Spatial Propagation Network (NLSPN)~\cite{Park2020NonLocalSP} for refinement. NLSPN's advantage lies in its ability to search for neighbors beyond adjacent nodes, effectively capturing long-range spatial dependencies. 
The effectiveness of NLSPN greatly depends  on the accuracy of estimated affinity.
To ensure stability during propagation, normalizing the affinities is essential. Incorporating the confidence of the initial depth prediction during normalization helps mitigate the adverse effects of unreliable depth values during propagation, as demonstrated in the following equation:
\begin{equation}
w_{m,n}^{i,j}=c^{i,j}\cdot\tanh(\hat{w}_{m,n}^{i,j})/\gamma 
\label{eq:affinity normalization}
\end{equation}
\begin{equation}
c=f_{\psi}(\widetilde{F}_{fused},M_{fused})
\label{eq:confidence head}
\end{equation}
$w_{m,n}^{i,j}$ and $\hat{w}_{m,n}^{i,j}$ denotes the normalized and estimated affinity of pixel (i,j) respectively. $c^{i,j}\in[0,1]$ represents it's confidence level, while $\gamma$ denotes the learnable normalization parameter. The Eq.~\ref{eq:affinity normalization} ensures that the cumulative sum of normalized affinities within the neighborhood remains bounded by one. Moreover, Eq.~\ref{eq:confidence head} reveals the optimized generation process of confidence map $c$, where $f_\psi(\cdot)$ denotes the confidence head which leverages attention mechanism to integrate temporal feature $\widetilde{F}_{fused}$ and fused masks ${M}_{fused}$. The fused masks are essentially similar to the confidence map, as they reflect the consensual spatial reliability of modalities, providing extra guidance in the learning of Eq.~\ref{eq:confidence head}.

Our RDR module significantly improves the effectiveness of NLSPN by enhancing the accuracy of estimated affinity and confidence.
In affinity estimation, the learned feature $\widetilde{F}_{fused}$ effectively aggregates spatial-temporal information, benefiting from the global horizon of attention layer and the convGRU layer's exploration of temporal correlation.
Regarding confidence estimation, the learned fused masks serve as informative extra features to enhance confidence map learning. Consequently, the RDR module with enhanced affinity enables spatial propagation to yield higher accuracy and fine-grained depth maps..

\subsection{Loss Functions} 

Our SRFNet is trained in a supervised manner, where the depth ground truth is captured from the LiDAR. Following RAMNet~\cite{Gehrig2021CombiningEA}, the depth values are normalized into log formats $\bar{\widehat{D}}_{k}\in[0,1]$, to facilitate predicting depths with large variations. The normalization can be formulated as follows:
\begin{equation}
\label{eq: log depth}
    \widehat{D}_{k}=\frac{1}{\alpha}\log\frac{\widehat{D}_{m,k}}{D_{\mathrm{max}}}+1,
\end{equation}
where $\alpha$ is a predefined parameter to map the closestvalid depth to 0. For MVSEC dataset, $\alpha=3.7$ and $D_{max}=80$m. 
Our loss comprises two components: MSE loss and multi-scale scale-invariant gradient matching loss, computed for valid ground truth labels and summed over a sequence of events and frames:
\begin{equation}
    L_{total}=\sum_{k=0}^{L-1}L_{k,mse}+\lambda{L_{k,grad}}
\end{equation}
where $L$ denotes the length of sequence fed for training. And the $\lambda$ is set to 0.25 in this paper. These losses are calculated based on the log depth difference $R_{k}=\widehat{D_{k}}-D_{k}$, where $\widehat{D_{k}}$ is the predicted depth and $D_{k}$ is the ground truth depth. The MSE loss and gradient matching loss can be formulated as:
\begin{equation}
    {L}_{k,mse}=\frac1n\sum_\mathbf{u}(R_k(\mathbf{u}))^2
\end{equation}
\begin{equation}
    {L}_{k,grad}=\frac{1}{n}\sum_{s}\sum_{\mathbf{u}}|\nabla_{x}R_{k}^{s}(\mathbf{u})|+|\nabla_{y}R_{k}^{s}(\mathbf{u})|
\end{equation}
Among them, $u$ indexes the valid ground truth pixels and $R^s_k(u)$ refers to the residual at scale $s$. $\nabla_{x}$ and $\nabla_{y}$ compute the edges in the x and y direction respectively by using the Sobel operator.

\begin{figure*}[t]
    \centering
    \includegraphics[width=0.93\textwidth]{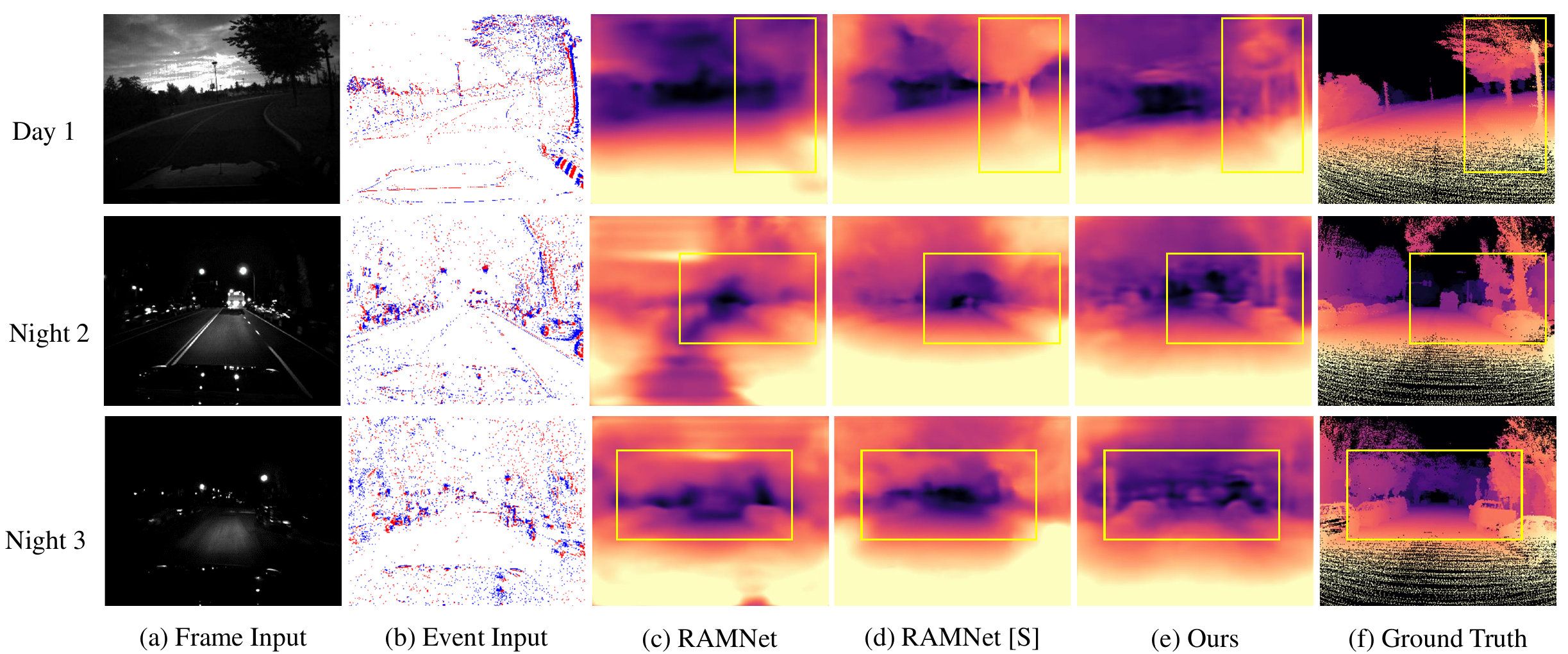}
    \vspace{-10pt}
    \caption{\textbf{Qualitative comparisons with different methods for MVSEC dataset}. (c) RAMNet is purely trained on MVSEC; (d) RAMNet [S] denotes RAMNet pre-trained on synthetic dataset. The yellow bounding box indicates the region of significant contrast.}
    \label{visualization comparison}
    \vspace{-6pt}
\end{figure*}

\begin{figure*}[t]
    \centering
    \includegraphics[width=0.93\textwidth]{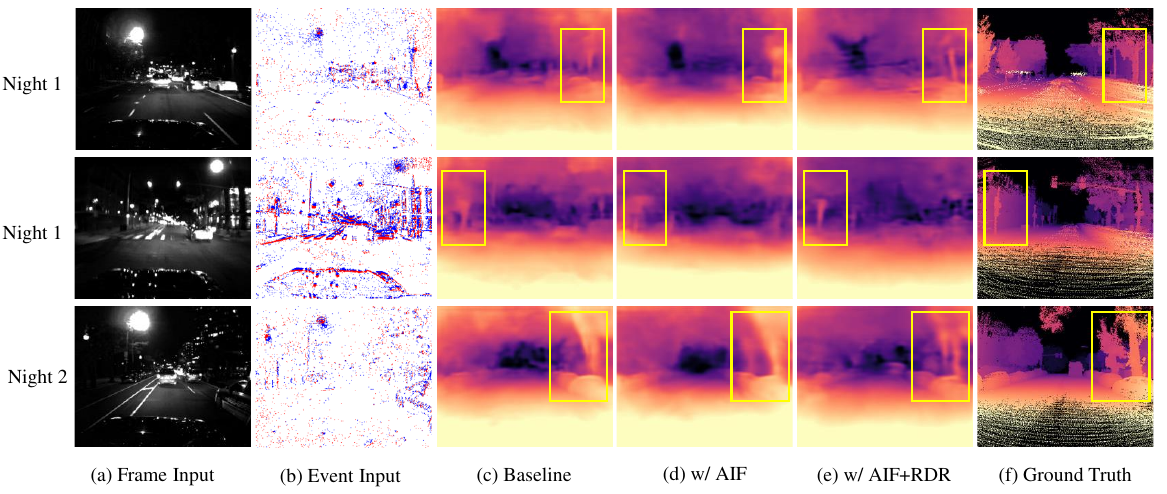}
    \vspace{-10pt}
    \caption{\textbf{Qualitative results of ablation studies of SRFNet on the MVSEC dataset}. (c) denotes the baseline; (d) is our SRFNet without the RDR module; (e) is the complete SRFNet.}
    \label{visualization ablation}
    \vspace{-6pt}
\end{figure*}

\begin{table*}[t]
\centering
\caption{Performance evaluation across multiple maximum cut-off depths on MVSEC Using Average Error (Lower is better). Notably, the best results are highlighted in bold, while the second-best outcomes are underlined. Our method excels in both daytime and nighttime scenarios among methods exclusively trained on real-world datasets. Furthermore, [S] means the methods are pre-trained on synthetic datasets.}
\label{tab:whole_comparison}
\vspace{-7pt}
\scalebox{0.86}{
\begin{tabular}{cc|cc|c|cccc}
\hline
        &          & \multicolumn{2}{c|}{Frame-based} & Event-based    & \multicolumn{4}{c}{Event and frame fusion}                             \\
        \hline
Dataset & Distance & MegaDepth~\cite{Li2018MegaDepthLS}       & MonoDEVS~\cite{Gurram2021MonocularDE}       & E2Dpeth{[}S{]}~\cite{HidalgoCarrio2020LearningMD} & RAMNet{[}S{]}~\cite{Gehrig2021CombiningEA} & RAMNet & Ours w/ AIF & Ours w/ AIF+RDR \\ \hline
\multirow{3}{*}{outdoor day1}   & 10m & 2.38 & 1.47 & 1.02 & 1.05 & 1.29 & \underline{0.97} & \textbf{0.96} \\
                                & 20m & 3.63 & 2.49 & 1.93 & 1.99 & 2.23 & \textbf{1.73} & \underline{1.77} \\
                                & 30m & 4.54 & 3.13 & 2.47 & 2.63 & 2.89 & \textbf{2.2}  & \underline{2.37} \\ \hline
\multirow{3}{*}{outdoor night1} & 10m & 4.64 & 2.99 & 1.78 & 2.31 & 6.31 & \underline{1.47} & \textbf{1.26} \\
                                & 20m & 7.55 & 3.71 & 2.45 & 2.92 & 6.03  & \underline{2.08} & \textbf{1.95} \\
                                & 30m & 8.8  & 5.08 & 3.26 & 3.64 & 6.71  & \underline{3.04} & \textbf{3.01} \\ \hline
\multirow{3}{*}{outdoor night2} & 10m & 4.28 & 1.77 & \textbf{1.09} & 1.11 & \underline{1.09} & 1.26 & 1.19 \\
                                & 20m & 7.04 & 3.17 & 2.20 & 2.26 & 2.93 & \textbf{2.08} & \underline{2.13} \\
                                & 30m & 8.39 & 4.66 & 3.36 & 3.49 & 4.26 & \textbf{3.25} & \underline{3.22} \\ \hline
\multirow{3}{*}{outdoor night3} & 10m & 3.71 & 1.40 & \textbf{0.75} & \underline{0.88} & 0.91 & 1.07 & 1.01 \\
                                & 20m & 6.50 & 3.01 & \textbf{1.91} & 2.31 & 2.35 & \underline{1.98} & 2.12  \\
                                & 30m & 7.79 & 4.68 & \textbf{3.03} & 3.79 & 3.95 & \underline{3.34} & 3.52 \\ \hline
\end{tabular}}
\end{table*}

\begin{table*}[]
\centering
\caption{The comparison with pre-train and retrained RAMNet under more metrics. $\downarrow$ denotes lower is better and $\uparrow$ denotes higher is better. Even without the help of the RDR module, our method still consistently outperforms pre-trained RAMNet or demonstrates comparable performance with inputs under different resolutions}
\vspace{-7pt}
\label{tab:detailed-comparison}
\scalebox{0.91}{
\begin{tabular}{l|c|c|cccccccc}
\hline
Resolution &
  Dataset &
  Methods &
  Abs Rel $\downarrow$ &
  Sq Rel $\downarrow$ &
  RMSE $\downarrow$ &
  RMSE log $\downarrow$ &
  SI log $\downarrow$ &
  $\delta<1.25\uparrow$ &
  $\delta<1.25^{2}\uparrow$ &
  $\delta<1.25^{3}\uparrow$ \\ \hline
\multirow{6}{*}{$224\times224$} &
  \multirow{3}{*}{Outdoor day1} &
  RAMNet [S] &
  0.246 &
  0.162 &
  \textbf{8.926} &
  0.378 &
  \textbf{0.072} &
  0.593 &
  0.806 &
  0.899 \\
 &
   &
  RAMNet &
  0.286 &
  0.233 &
  9.846 &
  0.452 &
  0.124 &
  0.587 &
  0.775 &
  0.865 \\
 &
   &
  Ours &
  \textbf{0.234} &
  \textbf{0.160} &
  9.150 &
  \textbf{0.364} &
  0.073 &
  \textbf{0.634} &
  \textbf{0.814} &
  \textbf{0.922} \\ \cline{2-11} 
 &
  \multirow{3}{*}{Outdoor night1} &
  RAMNet [S] &
  0.431 &
  0.739 &
  \textbf{11.822} &
  \textbf{0.530} &
  \textbf{0.135} &
  \textbf{0.477} &
  0.664 &
  \textbf{0.789} \\
 &
   &
  RAMNet &
  1.018 &
  3.447 &
  15.383 &
  0.831 &
  0.213 &
  0.265 &
  0.427 &
  0.560 \\
 &
   &
  Ours &
  \textbf{0.335} &
  \textbf{0.259} &
  13.608 &
  0.544 &
  0.143 &
  0.465 &
  \textbf{0.667} &
  0.787 \\ \hline
\multirow{6}{*}{$256\times320$} &
  \multirow{3}{*}{Outdoor day1} &
  RAMNet [S] &
  0.278 &
  \textbf{0.207} &
  8.820 &
  0.429 &
  0.094 &
  0.552 &
  0.772 &
  0.874 \\
 &
   &
  RAMNet &
  0.335 &
  0.364 &
  9.498 &
  0.507 &
  0.158 &
  0.552 &
  0.747 &
  0.842 \\
 &
   &
  Ours &
  \textbf{0.268} &
  0.244 &
  \textbf{8.453} &
  \textbf{0.375} &
  \textbf{0.079} &
  \textbf{0.637} &
  \textbf{0.810} &
  \textbf{0.900} \\ \cline{2-11} 
 &
  \multirow{3}{*}{Outdoor night1} &
  RAMNet [S] &
  0.445 &
  0.693 &
  \textbf{10.408} &
  0.534 &
  0.128 &
  \textbf{0.434} &
  0.652 &
  0.790 \\
 &
   &
  RAMNet &
  1.011 &
  3.490 &
  13.222 &
  0.808 &
  0.236 &
  0.291 &
  0.466 &
  0.597 \\
 &
   &
  Ours &
  \textbf{0.371} &
  \textbf{0.316} &
  11.469 &
  \textbf{0.521} &
  \textbf{0.120} &
  0.433 &
  \textbf{0.662} &
  \textbf{0.800} \\ \hline
\end{tabular}}
\end{table*}

\section{EXPERIMENTAL EVALUATION}

\subsection{Dataset and Implementation Details}
We conduct experiments on two datasets, \ie, MVSEC dataset and DENSE dataset. The MVSEC dataset is a real-world event dataset~\cite{Zhu2018TheMS}, while the DENSE dataset is a synthetic dataset~\cite{HidalgoCarrio2020LearningMD}. 

\noindent \textbf{MVSEC dataset:} It is recorded with a pair of DAVIS cameras. The resolution of frames and events is $346 \times 260$. For supervised training, the depth ground truth is obtained via a LiDAR scanner. This dataset covers not only daytime and nighttime driving sequences, but also indoor sequences captured by a quadcopter. We choose the subset "outdoor day2" for training and "day1", "night1", "night2" and "night3" for testing. The frames are recorded with grayscale. The depth maps, daytime and nighttime frames are recorded at 20Hz, 10Hz, and 45 Hz, respectively. 

\noindent \textbf{Implementation Details:} For event streams, we divide each voxel grid into 5 time bins. Similar with RAMNet~\cite{Gehrig2021CombiningEA}, we set the length of input sequences with paired events and frames to 8. We utilize Adam optimizer~\cite{kingma2014adam}. The initial learning rates set for the AIF module and RDR module are 5e-6 and 1e-5, respectively. The batch size is 4, and the number of training epoches is 100. For data augmentation, we employ data normalization, randomly cropping, and horizontal flipping.

\subsection{Evaluation on MVSEC and DENSE Datasets}
Previous methods~\cite{Gehrig2021CombiningEA,HidalgoCarrio2020LearningMD,Shi2023ImprovedED} are pre-trained on synthetic datasets. Considering that the amount of these synthetic datasets is huge, we skip the pre-training to avoid tedious training process. For adequate evaluation, we not only compare with the methods based on frames~\cite{Gurram2021MonocularDE, Li2018MegaDepthLS}, events~\cite{HidalgoCarrio2020LearningMD}, but also compare with the methods combining frames with events~\cite{Gehrig2021CombiningEA}. In addition, to evaluate our SRFNet and compared methods in the same training setting, we re-train MonoDEVS~\cite{Gurram2021MonocularDE} and RAMNet~\cite{Gehrig2021CombiningEA} on the MVSEC dataset. 


\begin{table*}[]
\centering
\caption{Ablation Study Conducted on the MVSEC Dataset to evaluate the different settings SRFNet by Using Average Error. The Best Results, Highlighted in Bold, Showcase the Effectiveness of each component of SRFNet Across All Analyzed Metrics}
\vspace{-7pt}
\label{tab:ablation-study}
\scalebox{0.95}{
\begin{tabular}{c|cc|cc|ccc|ccc|ccc|ccc}
\hline
 &
  \multicolumn{2}{c|}{AIF} &
  \multicolumn{2}{c|}{RDR} &
  \multicolumn{3}{c|}{day1} &
  \multicolumn{3}{c|}{night1} &
  \multicolumn{3}{c|}{night2} &
  \multicolumn{3}{c}{night3} \\ \cline{2-17} 
 &
  Attention &
  Mask &
  SPN &
  Mask &
  10m &
  20m &
  30m &
  10m &
  20m &
  30m &
  10m &
  20m &
  30m &
  10m &
  20m &
  30m \\ \hline
Baseline &
   &
   &
   &
   &
  \multicolumn{1}{c}{1.11} &
  \multicolumn{1}{c}{2.04} &
  2.57 &
  \multicolumn{1}{c}{1.93} &
  \multicolumn{1}{c}{2.48} &
  3.49 &
  \multicolumn{1}{c}{1.24} &
  \multicolumn{1}{c}{2.25} &
  3.46 &
  \multicolumn{1}{c}{\textbf{1.00}} &
  \multicolumn{1}{c}{2.24} &
  3.71 \\ \hline
(1) &
  \checkmark &
    &
    &
    &
  \multicolumn{1}{c}{1.07} &
  \multicolumn{1}{c}{1.85} &
  2.29 &
  \multicolumn{1}{c}{2.11} &
  \multicolumn{1}{c}{2.51} &
  3.58 &
  \multicolumn{1}{c}{1.72} &
  \multicolumn{1}{c}{2.44} &
  3.58 &
  \multicolumn{1}{c}{1.53} &
  \multicolumn{1}{c}{2.34} &
  3.59 \\
(2) &
  \checkmark &
  \checkmark &
    &
    &
  \multicolumn{1}{c}{0.97} &
  \multicolumn{1}{c}{\textbf{1.73}} &
  \textbf{2.20} &
  \multicolumn{1}{c}{1.3} &
  \multicolumn{1}{c}{1.97} &
  3.07 &
  \multicolumn{1}{c}{1.26} &
  \multicolumn{1}{c}{\textbf{2.08}} &
  \textbf{3.25} &
  \multicolumn{1}{c}{1.07} &
  \multicolumn{1}{c}{\textbf{1.98}} &
  \textbf{3.34} \\
(3) &
  \checkmark &
  \checkmark &
  \checkmark &
    &
  \multicolumn{1}{c}{1.07} &
  \multicolumn{1}{c}{1.79} &
  2.35 &
  \multicolumn{1}{c}{1.63} &
  \multicolumn{1}{c}{2.33} &
  3.48 &
  \multicolumn{1}{c}{1.26} &
  \multicolumn{1}{c}{2.33} &
  3.67 &
  \multicolumn{1}{c}{1.07} &
  \multicolumn{1}{c}{2.27} &
  3.81 \\
(4) &
  \checkmark &
  \checkmark &
  \checkmark &
  \checkmark &
  \multicolumn{1}{c}{\textbf{0.96}} &
  \multicolumn{1}{c}{1.77} &
  2.37 &
  \multicolumn{1}{c}{\textbf{1.26}} &
  \multicolumn{1}{c}{\textbf{1.95}} &
  \textbf{3.01} &
  \multicolumn{1}{c}{\textbf{1.19}} &
  \multicolumn{1}{c}{2.13} &
  3.32 &
  \multicolumn{1}{c}{1.01} &
  \multicolumn{1}{c}{2.12} &
  3.52 \\\hline
\end{tabular}
}
\vspace{-5pt}
\end{table*}


\noindent \textbf{Metrics.} Similar with RAMNet~\cite{Gehrig2021CombiningEA}, we choose average absolute depth
errors at different cut-off depth distances (i.e., 10m, 20m
and 30m) for comparison. The metrics we utilize include absolute relative error (Abs Rel), logarithmic mean squared error (RMSE log), scale invariant logarithmic error (SI log), and
accuracy ($\delta < 1.25^{n}, n = 1, 2, 3$) for detailed comparison with RAMNet.

\begin{figure}[t]
    \centering
    \includegraphics[width=0.8\linewidth]{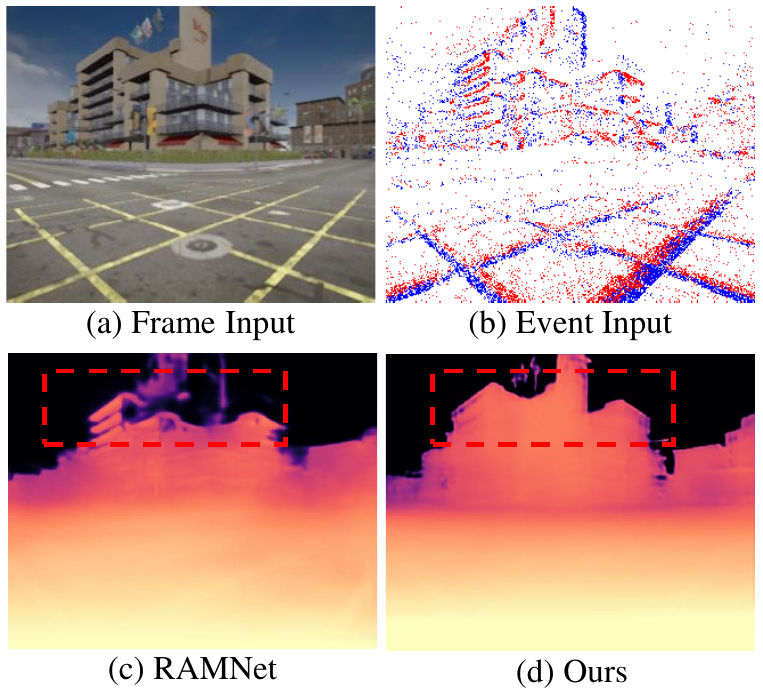}
    \caption{\textbf{Qualitative comparison with RAMNet on the DENSE dataset}; (c) is RAMNet trained on DENSE dataset; (d) is our SRFNet trained on DENSE dataset.}
    \label{synthetic visualization ablation}
    \vspace{-8pt}
\end{figure}

\begin{table}[]
\centering
\caption{Comparison with RAMNet on synthetic DENSE dataset~\cite{HidalgoCarrio2020LearningMD}.}
\vspace{-8pt}
\label{tab:synthetic comparison}
\begin{tabular}{c|ccc|cc}
\hline
\multirow{2}{*}{Method} & \multicolumn{3}{c|}{Avg. Error $\downarrow$}                  &                &                \\ \cline{2-6} 
                        & 10m            & 20m            & 30m            & Abs. Rel $\downarrow$      & RMSE log $\downarrow $     \\ \hline
RAMNet                  & 2.619          & 11.264         & 19.113         & 1.189          & 0.832          \\
Ours                    & \textbf{1.503} & \textbf{3.566} & \textbf{6.116} & \textbf{0.513} & \textbf{0.687} \\ \hline
\end{tabular}
\vspace{-10pt}
\end{table}
\noindent \textbf{Quantitative Evaluation.} As shown in Tab.~\ref{tab:whole_comparison}, we conduct comprehensive comparisons among different types of monocular depth estimation methods on various scenes. It can be seen that our method outperforms methods with single modality as input~\cite{Gurram2021MonocularDE, Li2018MegaDepthLS, HidalgoCarrio2020LearningMD} in most scenes. Note that our method has lower performance compared with E2Depth in some scenes, due to its pre-training on synthetic datasets. Moreover, compared with RAMNet that combines frames with events, our method even outperforms it even with pre-training. For example, compared with RAMNet with pre-training, our method obtains 0.22 gain with 20m distance in the outdoor day1 dataset. \textit{The results can also demonstrate the superior generalization of our method, as the training dataset we utilize contains no night scene.} 


In Tab.~\ref{tab:detailed-comparison}, we also provide a more detailed comparison with RAMNet, as it is the only open-source monocular depth estimation method that combines frames with events. Our SRFNet shows consistent superiority compared with RAMNet without pre-training in all input resolutions, sub-datasets and metrics. We also achieve competitive results with pre-trained RAMNet. For example, with $224 \times 224$ input resolution in the outdoor day1 sub-dataset, our method obtains 0.014 gain in the RMSE log metric. As shown in Fig.~\ref{visualization comparison}, our method predicts clearer trees in the first and third rows. Instead, RAMNet only predicts blurry structures on the trees. We ascribe it to the effectiveness of our proposed iterative feature fusion between frames and events.

We also conduct experiments on the synthetic DENSE dataset~\cite{HidalgoCarrio2020LearningMD}, following RAMNet~\cite{Gehrig2021CombiningEA}, We set the $\alpha=5.7$ and $D_{\mathrm{max}}=1000$ in Eq. \ref{eq: log depth} for DENSE. As the quantitative comparison shown in Tab.~\ref{tab:synthetic comparison}, our SRFNet outperforms RAMNet under all of evaluation metrics. 
Additionally the visualization result comparison is depicted in Fig.~\ref{synthetic visualization ablation}, our SRFNet achieves more fine-grained prediction compared with RAMNet.


\subsection{Ablation Study}

\noindent \textbf{AIF module.} The proposed AIF module is based on the attention mechanism to interactively update the fused features and fused masks. To verify its effectiveness, we first add the attention mechanism into AIF. The performance has an improvement in most scenes. However, in some scenes such as "night3", the performance drops as the direct fusion neglects the structural ambiguity between two modalities. Furthermore, as shown in Tab.\ref{tab:ablation-study}, by adding the spatial priors, \ie, the mask learning, into the AIF module, the performance has an obvious improvement, \eg, 0.12 gain in 20m metric on day1 sub-dataset. It demonstrates that learning the spatial reliability is powerful for the fusion of frames and events.

\noindent \textbf{RDR module.} To obtain the fine-grained depth estimation, we further add the RDR module. The RDR module is based on the SPN, whose confidence map is refined by the fused masks from the AIF module. In this case, we first simply employ the original SPN, and find that it less effective. We ascribe it that the direct predicting the relationships between pixels in the SPN is difficult, especially when the two modalities exhibit different spatial priors. Instead, by integrating the fused masks with the SPN, the performance can be improved such as in the night1 sub-dataset. As shown in Fig.~\ref{visualization ablation}, it can be found that by adding AIF module and RDR module, the trees in the three rows have clearer structures. 



\section{CONCLUSION}
In this paper, we proposed a novel monocular depth estimation network, called SRFNet, which can effectively fuse the frames and events by considering the spatial reliability. This was based on the observation that frames and events might drop their reliability in the challenging conditions, such as high-speed motion scenes for the frames and scenes with marginal light change for the events. Therefore, direct fusing the two modalities would cause structural ambiguity and degrade the depth estimation performance. Accordingly, we proposed the AIF module to learn the consensus regions of the two modalities, which can guide the inter-modal feature fusion and enhance each modality's feature learning process. In addition, we proposed the RDR module to obtain fine-grained depth estimation by integrating the fused spatial priors from the AIF module. 
We demonstrated that our SRFNet outperforms existing SOTA methods in various scenes. Detailed ablation studies also show the superiority of the AIF module and RDR module, especially about the mask learning strategy.

\noindent \textbf{Future work:} We plan to generalize our fusion method to more tasks. We hope our thinking of the fusion technique can inspire the community for better combining the novel event sensors with traditional RGB cameras.

\noindent \textbf{Acknowledgement:} This paper is supported by the National Natural Science Foundation of China (NSF) under Grant No. NSFC22FYT45 and the Guangzhou City, University and Enterprise Joint Fund under Grant No.SL2022A03J01278

\clearpage
{\small
\bibliographystyle{ieee_fullname}
\bibliography{egbib}
}

\end{document}


\begin{table*}[]
\centering
\caption{Detailed Comparison of Reliability Mask Usage in the RDR Module. Evaluation at a Resolution of ($256\times320$) Input, Using the MVSEC Dataset.}
\label{tab:my-table}
\begin{tabular}{c|c|cccccccc}
\hline
Dataset                         & RDR   & Abs Rel$\downarrow$ & Sq Rel$\downarrow$ & RMSE$\downarrow$   & RMSE log$\downarrow$ & SI log$\downarrow$      &    $\delta<1.25\uparrow$ &
  $\delta<1.25^{2}\uparrow$ &
  $\delta<1.25^{3}\uparrow$ \\ \hline
\multirow{2}{*}{Outdoor day1}   & SPN & 0.309   & 0.290   & 8.616  & 0.396    & \textbf{0.078} & 0.547 & 0.795 & 0.896 \\
 & Mask + SPN & \textbf{0.268} & \textbf{0.243} & \textbf{8.453}  & \textbf{0.375} & 0.079         & \textbf{0.637} & \textbf{0.810}  & \textbf{0.900} \\ \hline
\multirow{2}{*}{Outdoor night1} & SPN & 0.499   & 0.516  & 11.625 & 0.588    & 0.121          & 0.299 & 0.548 & 0.736 \\
 & Mask + SPN & \textbf{0.371} & \textbf{0.316} & \textbf{11.469} & \textbf{0.521} & \textbf{0.120} & \textbf{0.433} & \textbf{0.662} & \textbf{0.800} \\ \hline
\end{tabular}
\end{table*}

\begin{table*}[]
\centering
\caption{Detailed Comparison with RAMNet on Synthetic Dataset DENSE. The Input Resolution is ($256\times320$).}
\label{tab:my-table}
\resizebox{\textwidth}{13mm}{
\begin{tabular}{c|c|c|cccccccc|ccc}
\hline
 &
  \multirow{2}{*}{Dataset} &
  \multirow{2}{*}{Methods} &
  \multirow{2}{*}{Abs. Rel$\downarrow$} &
  \multirow{2}{*}{Sq. Rel$\downarrow$} &
  \multirow{2}{*}{RMSE$\downarrow$} &
  \multirow{2}{*}{RMSE log$\downarrow$} &
  \multirow{2}{*}{SI log$\downarrow$} &
  \multirow{2}{*}{$\delta<1.25\uparrow$} &
  \multirow{2}{*}{$\delta<1.25^{2}\uparrow$} &
  \multicolumn{1}{c|}{\multirow{2}{*}{$\delta<1.25^{3}\uparrow$}} &
  \multicolumn{3}{c}{Avg. Error$\downarrow$} \\
 &
   &
   &
   &
   &
   &
   &
   &
   &
   &
  \multicolumn{1}{c|}{} &
  10m &
  20m &
  30m \\ \hline
\multirow{4}{*}{Validation sets} &
  \multicolumn{1}{c|}{\multirow{2}{*}{Town 06}} &
  \multicolumn{1}{c|}{RAMNet} &
  0.331 &
  8.615 &
  102.116 &
  0.416 &
  0.138 &
  0.707 &
  0.866 &
  0.939 &
  1.531 &
  4.49 &
  6.549 \\
 &
  \multicolumn{1}{c|}{} &
  \multicolumn{1}{c|}{Ours} &
  \textbf{0.21} &
  \textbf{5.099} &
  \textbf{80.941} &
  \textbf{0.308} &
  \textbf{0.087} &
  \textbf{0.881} &
  \textbf{0.954} &
  \textbf{0.974} &
  \textbf{0.854} &
  \textbf{2.756} &
  \textbf{4.033} \\ \cline{2-14} 
 &
  \multicolumn{1}{c|}{\multirow{2}{*}{Town 07}} &
  \multicolumn{1}{c|}{RAMNet} &
  0.697 &
  20.121 &
  111.752 &
  0.573 &
  0.267 &
  0.704 &
  0.821 &
  0.885 &
  2.876 &
  9.042 &
  13.18 \\
 &
  \multicolumn{1}{c|}{} &
  \multicolumn{1}{c|}{Ours} &
  \textbf{0.26} &
  \textbf{5.481} &
  \textbf{72.416} &
  \textbf{0.371} &
  \textbf{0.11} &
  \textbf{0.77} &
  \textbf{0.886} &
  \textbf{0.936} &
  \textbf{1.321} &
  \textbf{3.505} &
  \textbf{4.647} \\ \hline
\multirow{2}{*}{Test set} &
  \multicolumn{1}{c|}{\multirow{2}{*}{Town 10}} &
  \multicolumn{1}{c|}{RAMNet} &
  1.189 &
  36.709 &
  198.728 &
  0.832 &
  0.54 &
  0.592 &
  \textbf{0.749} &
  \textbf{0.821} &
  2.62 &
  11.264 &
  19.113 \\
 &
  \multicolumn{1}{c|}{} &
  \multicolumn{1}{c|}{Ours} &
  \textbf{0.513} &
  \textbf{9.309} &
  \textbf{126.374} &
  \textbf{0.687} &
  \textbf{0.332} &
  \textbf{0.596} &
  0.742 &
  0.795 &
  \textbf{1.503} &
  \textbf{3.566} &
  \textbf{6.116} \\ \hline
\end{tabular}
}
\end{table*}